\documentclass[11pt,letterpaper]{article}
\usepackage{graphicx}
\usepackage{amsmath,amssymb,amsthm}
\usepackage[margin=1in]{geometry}

\usepackage{url}           
\usepackage{booktabs}     
\usepackage{amsfonts}      
\usepackage{nicefrac}      
\usepackage{microtype}    

\usepackage[algo2e,lined,vlined,boxed]{algorithm2e}
\usepackage{algorithm}

\usepackage{epstopdf}

\usepackage{caption}
\usepackage{subcaption}

\usepackage{hyperref}
\usepackage{color}
\usepackage{bbm}
\usepackage{multirow}

\usepackage[titletoc]{appendix}
\usepackage[export]{adjustbox}
\usepackage{wrapfig}
\usepackage{natbib}

\usepackage{authblk}

\newtheorem{lemma}{Lemma}

\newtheorem{theorem}{Theorem}

\newtheorem*{lemma*}{Lemma}
\newtheorem{condition}{Condition}

\def\real{\mathbb{R}}
\def\grad{\nabla}

\def\xvec{\textsc{\bf x}}

\def\prob{\mathbb{P}}

\def\supp{\text{supp}}

\newcommand{\norm}[2]{\left\lVert #1 \right\rVert_{#2}}

\DeclareMathOperator*{\argmin}{arg\,min}

\newcommand\bovermat[2]{%
  \makebox[0pt][l]{$\smash{\overbrace{\phantom{%
    \begin{matrix}#2\end{matrix}}}^{\text{#1}}}$}#2}

\title{Gradient Coding: Avoiding Stragglers in Synchronous Gradient Descent}

\author[$\star$]{Rashish Tandon}
\author[$\dagger$]{Qi Lei}
\author[$\ddagger$]{Alexandros G. Dimakis}
\author[$\sharp$]{Nikos Karampatziakis}
\affil[$\star$]{Department of Computer Science, UT Austin}
\affil[$\dagger$]{Institute for Computational Engineering and Sciences, UT Austin}
\affil[$\ddagger$]{Department of Electrical and Computer Engineering, UT Austin}
\affil[$\sharp$]{Microsoft}
\affil[ ]{\texttt{\{rashish@cs, leiqi@ices, dimakis@austin\}.utexas.edu, nikosk@microsoft.com}}

\date{\today}

\begin{document}
\maketitle
\begin{abstract} 
We propose a novel coding theoretic framework for mitigating stragglers in distributed learning. We show how carefully replicating data blocks and coding across gradients can provide tolerance to failures and stragglers for Synchronous Gradient Descent. We implement our schemes in python (using MPI) to run on Amazon EC2, and show how we compare against baseline approaches in running time and generalization error. 
\end{abstract} 
\section{Introduction}
We propose a novel coding theoretic framework for mitigating stragglers in distributed learning. 
The central idea can be seen through the simple example of Figure 1: Consider synchronous Gradient 
Descent (GD) on three workers ($W_1$,$W_2$,$W_3$). The baseline vanilla system is shown in the left figure and operates as follows: $\,$The three workers have different partitions of the labeled data stored locally ($D_1$,$D_2$,$D_3$) and all share the current model. Worker 1 computes the gradient of the model on examples in partition $D_1$, denoted by $g_1$. Similarly, Workers 2 and 3 compute $g_2$ and $g_3$. 
The three gradient vectors are then communicated to a central node (called the master/aggregator) $A$ which computes the full gradient by summing these vectors $g_1+g_2+g_3$ and updates the model with a gradient step. The new model is then sent to the workers and the system moves to the next round (where the same examples or other labeled examples, say $D_4$,$D_5$,$D_6$, will be used in the same way). 
\begin{figure}[h]
	\begin{center}
	\begin{subfigure}[b]{0.4\textwidth}
		\includegraphics[scale=0.25]{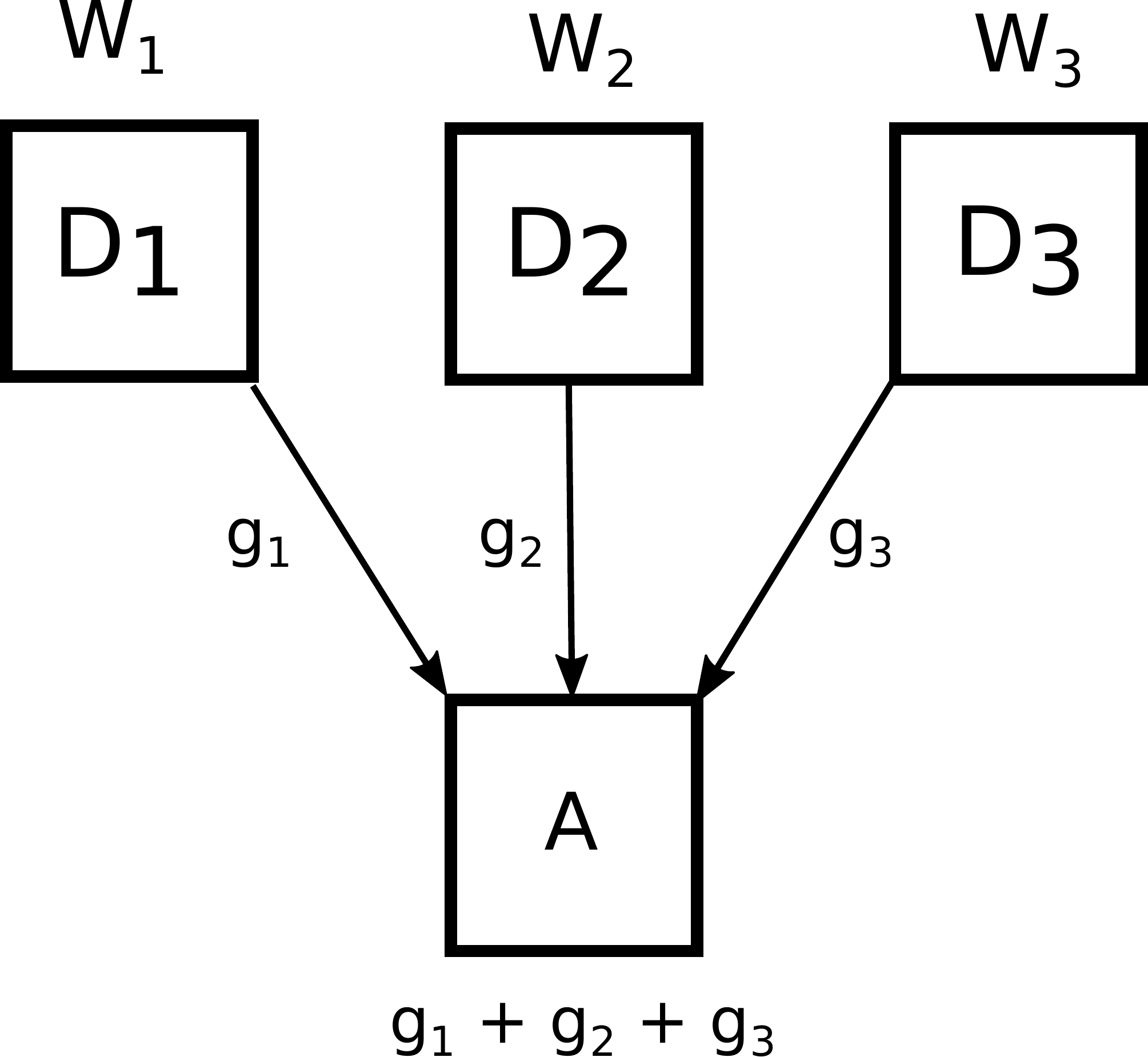}
		\vspace*{0.1mm}
		\caption{Naive synchronous gradient descent\vspace*{0.8cm}}
		\label{fig:naive}
	\end{subfigure}
	\begin{subfigure}[b]{0.4\textwidth}
		\includegraphics[scale=0.25]{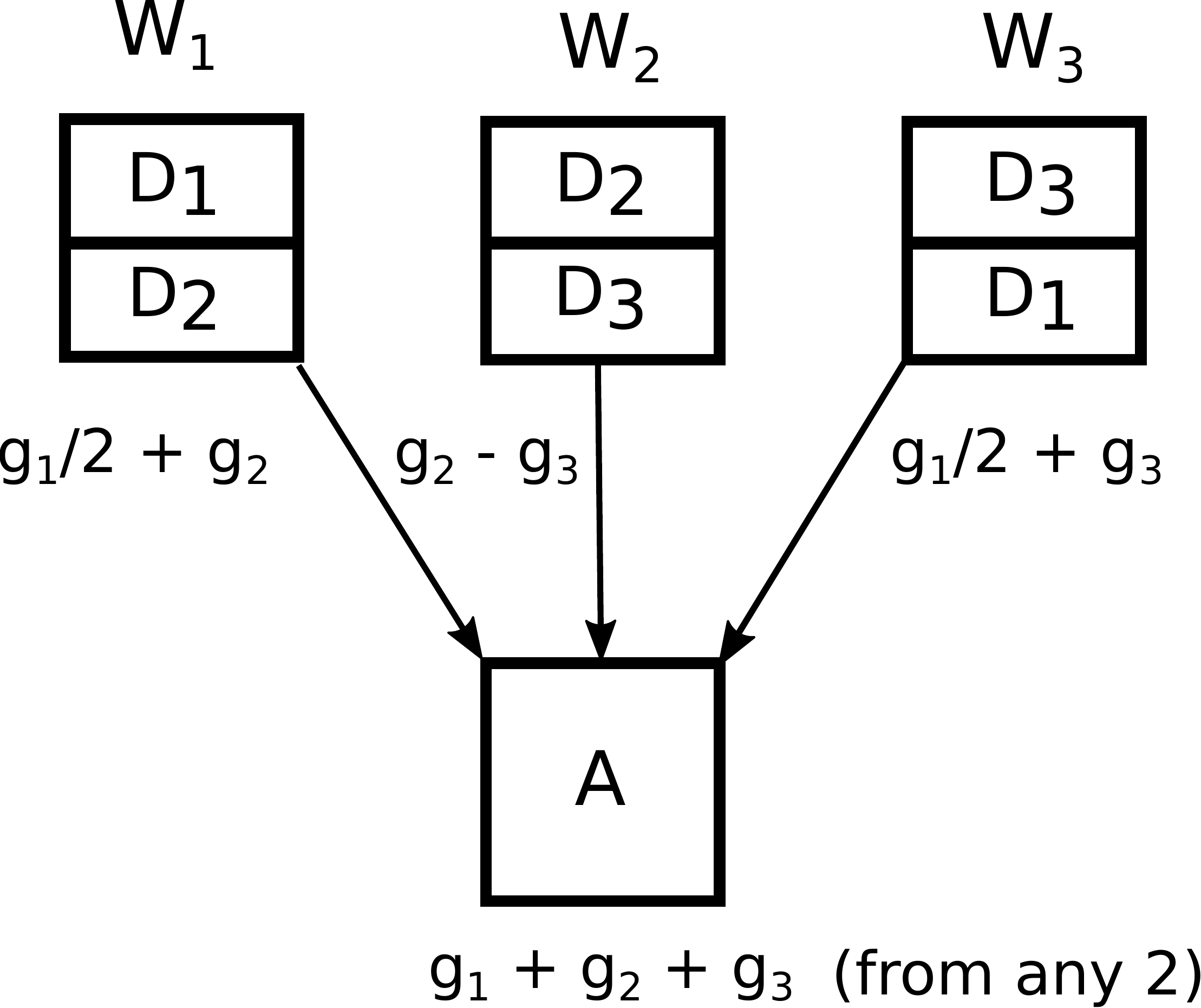}
		\vspace*{0.1mm}
		\caption{Gradient coding: The vector $g_1+g_2+g_3$ is in the span of \textit{any two} out of the vectors 
		$g_1/2 + g2$, $g_2-g_3$ and $g_1/2 + g_3$. }
		\label{fig:coded}
	\end{subfigure}
	\end{center}
	\caption{The idea of Gradient Coding.}
\end{figure}
The problem is that sometimes worker nodes can be stragglers~\citep{li2014communication, ho2013more, NIPS2012_4687} \textit{i.e.} delay significantly in computing and communicating gradient vectors to the master. This is especially pronounced for cheaper virtual machines in the cloud. For example on \texttt{t2.micro} machines on Amazon EC2, as can be seen in Figure~\ref{fig:commtime}: 
some machines can be $5\times$ slower in computing and communicating gradients compared to typical performance.\\ 

First, we discuss one way to resolve this problem if we replicate some data across machines by considering the placement in Fig.1 (b) but without coding. As can be seen, in Fig. 1 (b) each example is replicated two times using a specific placement policy. Each worker is assigned to compute two gradients on the two examples they have for this round. For example, $W_1$ will compute vectors $g_1$ and $g_2$. Now let's assume that $W_3$ is the straggler. If we use control messages, $W_1,W_2$ can notify the master $A$ that they are done. Subsequently, if feedback is used, the master can ask $W_1$ to send $g_1$ and $g_2$ and $W_2$ to send $g_3$. These feedback control messages can be much smaller than the actual gradient vectors but are still a system complication that can cause delays. However, feedback makes it possible for a centralized node to coordinate the workers, thereby avoiding stragglers. One can also reduce network communication further by simply asking $W_1$ to send the sum of two gradient vectors $g_1+g_2$ instead of sending both.  The master can then create the global gradient on this batch by summing these two vectors. Unfortunately, which linear combination must be sent depends on who is the straggler: If $W_2$ was the straggler then $W_1$ should be sending $g_2$ and $W_3$ sending $g_1+g_3$ so that their sum is the global gradient $g_1+g_2+g_3$.\\

In this paper we show that feedback and coordination is not necessary: every worker can send a \textit{single linear combination of gradient vectors} without knowing who the straggler will be. The main coding theoretic question we investigate is how to design these linear combinations so \textit{that any two} (or any fixed number generally) contain the $g_1+g_2+g_3$ vector in their span. In our example, in Fig. \ref{fig:coded},  $W_1$ sends $\frac{1}{2}g_1+ g_2$, $W_2$ sends $g_2-g_3$ and $W_3$ sends $\frac{1}{2}g_1+ g_3$. The reader can verify that $A$ can obtain the vector $g_1+g_2+g_3$ from \textit{any two out of these three vectors}. For instance, $g_1 + g_2 + g_3 = 2\left(\frac{1}{2}g_1+ g_2\right) - \left(g_2-g_3\right)$. We call this idea \textit{gradient coding}.\\

We consider this problem in the general setting of $n$ machines and \textbf{any} $s$ stragglers. We first establish a lower bound: to compute gradients on all the data in the presence of \textbf{any} $s$ stragglers, each partition must be replicated $s+1$ times across machines. We propose two placement and gradient coding schemes that match this optimal $s+1$ replication factor. We further consider a partial straggler setting, wherein we assume that a straggler can compute gradients at a fraction of the speed of others, and show how our scheme can be adapted to such scenarios. All proofs can be found in the appendix.\\

We also compare our scheme with the popular \textit{ignoring the stragglers} approach~\citep{2016arXiv160400981C}: simply doing a gradient step when most workers are done. We see that while ignoring the stragglers is faster, this loses some data which can hurt the generalization error. This can be especially pronounced in supervised learning with unbalanced labels or heavily unbalanced features since a few examples may contain critical, previously unseen information. 

\subsection{The Effects of Stragglers}
\begin{figure}[h]
\centering\hspace*{1in}
\begin{subfigure}{0.8\textwidth}
\includegraphics[scale=0.6]{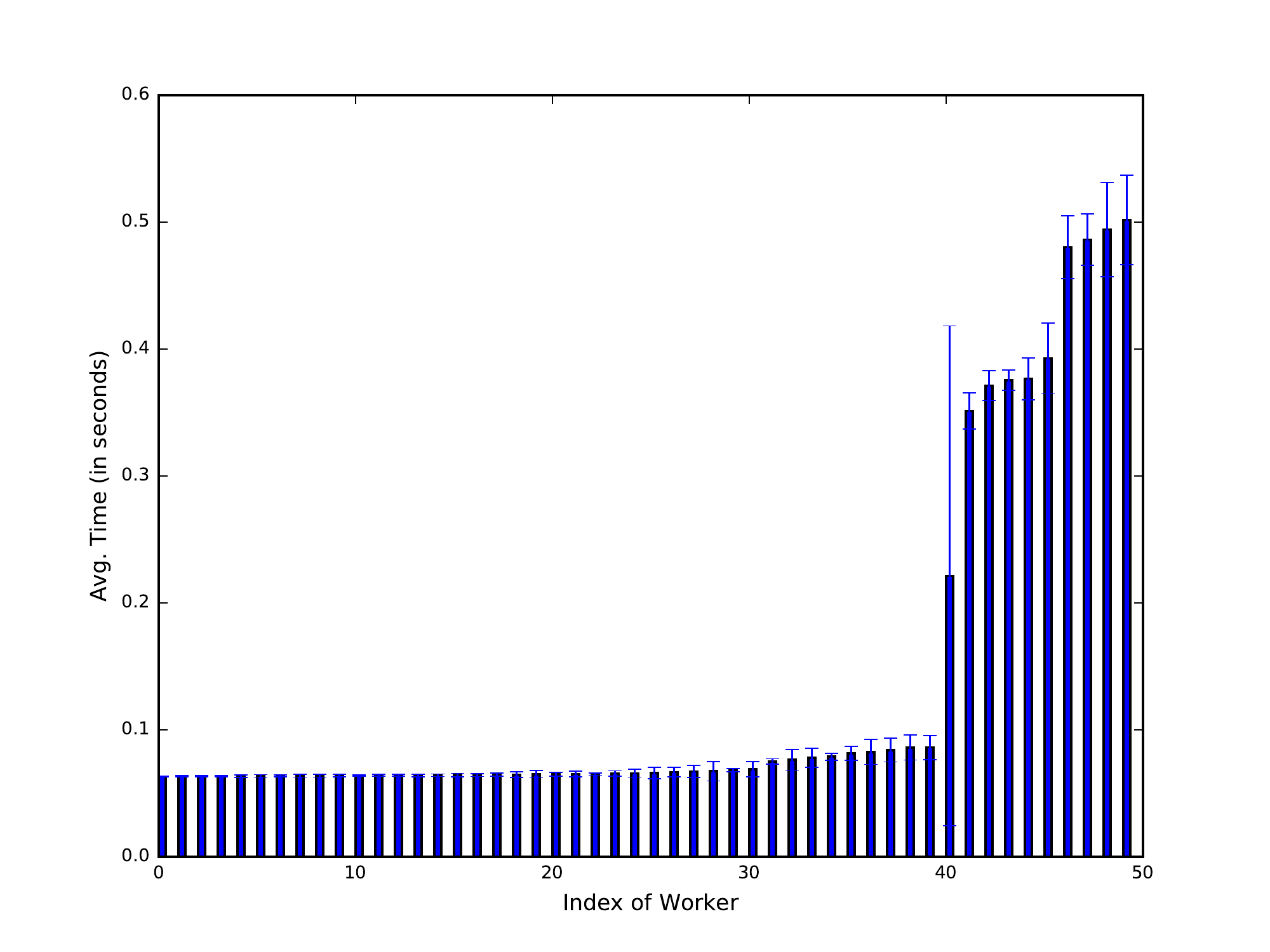}
\end{subfigure}
\caption{Average communication times, measure over $100$ rounds, for a vector of dimension $p=500000$ using $n=50$ \texttt{t2.micro} worker machines (and a \texttt{c3.8xlarge} master machine). Error bars indicate one standard deviation. }\label{fig:commtime}
\end{figure} 
In Figure \ref{fig:commtime}, we show the average time required for 50 \texttt{t2.micro} Amazon EC2 instances to communicate gradients to a single master machine (a \texttt{c3.8xlarge} instance). We observe that a few worker machines incurred a communication delay of up to $5\times$ the typical behavior. Interestingly, throughout the timescale of our experiments (a few hours), the straggling behavior was consistent in the same machines.\\

We have also experimented extensively with other Amazon EC2 instances: Our finding is that cheaper instance types have significantly higher variability in performance. This is especially true for \texttt{t2} type instance which on AWS are described as having \textit{Burstable Performance}. Fortunately, these machines have very low cost.\\

The choices of the number and type of workers used in training big models ultimately depends on total cost and time needed until deployment. The main message of this paper is that going for very low-cost instances and using coding to mitigate stragglers, may be a sensible choice for some learning problems.
\subsection{Related Work}
The slow machine problem is the Achilles heel of many distributed learning systems that run in modern cloud environments. Recognizing that, some recent work has advocated asynchronous approaches~\citep{li2014communication,ho2013more,ioannis16} to learning. While asynchronous updates are a valid way to avoid slow machines, they do give up many other desirable properties, including faster convergence rates, amenability to analysis, and ease of reproducibility and debugging.\\

Attacking the straggling problem in synchronous machine learning algorithms has surprisingly not received much attention in the literature. There do exist general systems solutions such as speculative execution~\cite{zaharia2008improving} but we believe that approaches tailored to machine learning can be vastly more efficient. In~\cite{2016arXiv160400981C} the authors use synchronous minibatch SGD and request a small number of additional worker machines so that they have an adequate minibatch size even when some machines are slow. However, this approach does not handle well machines that are consistently slow and the data on those machines might never participate in training. In~\cite{narayanamurthy2013} the authors describe an approach for dealing with failed machines by approximating the loss function in the failed partitions with a linear approximation at the last iterate before they failed. Since the linear approximation is only valid
at a small neighborhood of the model parameters, this approach can only work if failed data partitions are restored fairly quickly.\\

The work of~\cite{kangwook15} is the closest in spirit to our work, using coding theory and treating stragglers as erasures in the transmission of the computed results. However, we focus on codes for recovering the batch gradient of \textit{any} loss function while~\cite{kangwook15} and the more recent work of~\cite{grover16} describe techniques for mitigating stragglers in two different distributed applications: data shuffling and matrix multiplication. We also mention \cite{songze16}, which investigates a generalized view of the coding ideas in \cite{kangwook15}, showing that their solution is a single operating point in a general scheme of trading off latency of computation to the load of communication. Further closely related work has shown how coding can be used for distributed MapReduce, as well as a similar communication and computation tradeoff~\citep{limapred15,limapred16}. All these prior works develop novel coding techniques, but do not code across gradient vectors in the way we are proposing in this paper. 
\section{Preliminaries}
Given data $\mathbf{D} = \{(x_1, y_1),\ldots, (x_d,y_d)\}$, with each tuple $(x,y)\in \real^{p}\times \real$, several machine learning tasks aim to solve the following problem: 
\begin{equation}
\beta^* = \argmin_{\beta\in \real^p} \sum_{i=1}^{d} \ell\left(\beta; x_i,y_i\right) + \lambda R(\beta)
\end{equation}
where $\ell(\cdot)$ is a task-specific loss function, and $R(\cdot)$ is a regularization function. Typically, this optimization problem can be solved using gradient-based approaches. Let $g:= \sum_{i=1}^{d}\grad \ell(\beta^{(t)}; x_i, y_i)$ be the gradient of the loss at the current model $\beta^{(t)}$. Then the updates to the model are of the form: 
\begin{equation}
\beta^{(t+1)} = h_R\left(\beta^{(t)},g\right)
\end{equation}
where $h_R$ is a gradient-based optimizer, which also depends on $R(\cdot)$. Several methods such as gradient descent, accelerated gradient, conditional gradient (Frank-Wolfe), proximal methods, LBFGS, and bundle methods fit in this framework. However, if the number of samples, $d$, is large, a computational bottleneck in the above update step is the computation of the gradient, $g$, whose computation can be distributed.

\subsection{Notation}
Throughout this paper, we let $d$ denote the number of samples, $n$ denote the number of workers, $k$ denote the number of data partitions, and $s$ denote the number of stragglers/failures. The $n$ workers are denoted as $W_1, W_2, \ldots, W_n$. The partial gradients over $k$ data partitions are denoted as $g_1, g_2, \ldots, g_k$. The $i^{th}$ row of some matrices $A$ or $B$ is denoted as $a_i$ or $b_i$ respectively. For any vector $\xvec\in\real^n$, $\supp(\xvec)$ denotes its support \textit{i.e.} $\supp(\xvec) = \{i\,\vert\, \xvec_i\ne 0\}$, and $\norm{\xvec}{0}$ denotes its $\ell_0$-norm \text{i.e.} the cardinality of the support. $\mathbf{1}_{p\times q}$ and $\mathbf{0}_{p\times q}$ denote all $1$s and all $0$s matrices respectively, with dimension $p\times q$. Finally, for any $r\in\mathbb{N}$, $[r]$ denotes the set $\{1,\ldots, r\}$.

\subsection{The General Setup}
We can generalize the scheme in Figure \ref{fig:coded} to $n$ workers and $k$ data partitions by setting up a system of linear equations:
\begin{equation}\label{eq:abdecompdef}
AB = \mathbf{1}_{f\times k}
\end{equation}
where $f$ denotes the number of combinations of surviving workers/non-stragglers, $\mathbf{1}_{f\times k}$ is the all $1$s matrix of dimension $f\times k$, and we have matrices $A\in \real^{f\times n}$, $B\in \real^{n\times k}$.\\

We associate the $i^{th}$ row of $B$, $b_i$, with the $i^{th}$ worker, $W_i$. The support of $b_i$, $\supp(b_i)$, corresponds to the data partitions that worker $W_i$ has access to, and the entries of $b_i$ encode a linear combination over their gradients that worker $W_i$ transmits. Let $\bar{g} \in \real^{k\times d}$ be a matrix with each row being the partial gradient of a data partition \text{i.e.} 
\[
 \bar{g} = [g_1, g_2, \ldots, g_k]^T.
\]
Then, worker $W_i$ transmits $b_i \bar{g}$. Note that to transmit $b_i \bar{g}$, $W_i$ only needs to compute the partial gradients on the partitions in $\supp(b_i)$. Now, each row of $A$ is associated with a specific failure/straggler scenario, to which tolerance is desired. In particular, any row $a_i$, with support $\supp(a_i)$, corresponds to the scenario where the worker indices in $\supp(a_i)$ are alive/non-stragglers. Also, by the construction in Eq. \eqref{eq:abdecompdef}, we have:
\begin{align}
a_i B \bar{g} &= [1, 1, \ldots, 1] \bar{g} = \left(\sum_{j=1}^{k} g_j\right)^T \,\text{and,}\\
a_i B \bar{g} &= \sum_{k\in \supp(a_i)} a_i(k) (b_k \bar{g})
\end{align}
where $a_i(k)$ denotes the $k^{th}$ element of the row $a_i$. Thus, the entries of $a_i$ encode a linear combination which, when taken over the transmitted gradients of the alive/non-straggler workers, $\{b_k \bar{g}\}_{k\in \supp(a_i)}$, would yield the full gradient.\\

Going back to the example in Fig.~\ref{fig:coded}, the corresponding $A$ and $B$ matrices under the above generalization are: 
\begin{equation}
A = \begin{pmatrix}
0 & 1 & 2\\
1 & 0 & 1\\
2 & -1 & 0
\end{pmatrix}, \; \text{and}\;
B = \begin{pmatrix}
1/2 & 1 & 0\\
0 & 1 & -1\\
1/2 & 0 & 1
\end{pmatrix}
\end{equation}
with $f=3, n=3, k=3$. It is easy to check that $AB = \mathbf{1}_{3\times 3}$. Also, since every row of $A$ here has exactly one zero, we say that this scheme is robust to any one straggler.\\

In general, we shall seek schemes, through the construction of $(A,B)$, which are robust to \textbf{any} $s$ stragglers.\\

The rest of this paper is organized as follows. In Section \ref{sec:fullstragg} we provide two schemes applicable to any number of workers $n$, under the assumption that stragglers can be arbitrarily slow to the extent of total failure. In Section \ref{sec:partstragg}, we relax this assumption to the case of worker slowdown (with known slowdown factor), instead of failure, and show how our constructions can be appended to be more effective. Finally, in Section \ref{sec:expts} we present results of empirical tests using our proposed distribution schemes on Amazon EC2.
\section{Full Stragglers}\label{sec:fullstragg}
In this section, we consider schemes robust to any $s$ stragglers, given $n$ workers (with $s<n$). We assume that any straggler is (what we call) a \textit{full straggler} \textit{i.e.} it can be arbitrarily slow to the extent of complete failure. We show how to construct the matrices $A$ and $B$, with $AB = \mathbf{1}$, such that the scheme $(A,B)$ is robust to \textbf{any} $s$ full stragglers.\\

Consider any such scheme $(A,B)$. Since every row of $A$ represents a set of non-straggler workers, all possible sets over $[n]$ of size $(n-s)$ must be supports in the rows of $A$. Thus $f = \binom{n}{n-s} = \binom{n}{s}$ \textit{i.e.} the total number of failure scenarios is the number of ways to choose $s$ stragglers out of $n$ workers. Now, since each row of $A$ represents a linear span over some rows of $B$, and since we require $AB = \mathbf{1}$, this leads us to the following condition on $B$:
\begin{condition}[B-Span]\label{cond:bspan}
Consider any scheme $(A,B)$ robust to \textbf{any} $s$ stragglers, given $n$ workers (with $s<n$). Then we require that for every subset $I\subseteq [n], \lvert I\rvert = n-s$: 
\begin{equation}
\quad \mathbf{1}_{1\times k} \in \text{span}\{b_i \,\vert\, i\in I\}
\end{equation}
where $\text{span}\{\cdot\}$ is the span of vectors.
\end{condition}
The B-Span condition above ensures that the all $\mathbf{1}$s vector lies in the span of any $n-s$ rows of $B$. This is of course necessary. However, it is also sufficient. In particular, given a $B$ satisfying Condition~\ref{cond:bspan}, we can construct $A$ such that $AB=\mathbf{1}$, and $A$ has the support structure discussed above. The construction of $A$ is described in Algorithm~\ref{alg:geta} (in MATLAB syntax), and we have the following lemma.

\begin{lemma}\label{lem:getA}
Consider $B\in \real^{n\times k}$ satisfying Condition~\ref{cond:bspan} for some $s<n$. Then, Algorithm~\ref{alg:geta}, with input $B$ and $s$, yields an $A\in \real^{\binom{n}{s}\times n}$ such that $AB = \mathbf{1}_{\binom{n}{s}\times n}$ and the scheme $(A,B)$ is robust to any $s$ full stragglers.
\end{lemma}
\begin{algorithm}[b]
\SetAlgoLined
\SetKwInOut{Input}{Input}\SetKwInOut{Output}{Output}
\Input{$B$ satisfying Condition~\ref{cond:bspan}, $s (<n)$ }
\Output{$A$ such that $AB=\mathbf{1}_{\binom{n}{s}\times n}$}
\BlankLine
$f = \textrm{binom}(n,s);$\\
$A = \textrm{zeros}(f,n);$\\
\ForEach{$I\subseteq [n]\text{ s.t. } \lvert I\rvert = (n-s)$}{
$a = \textrm{zeros}(1,k);$\\
$x = \textrm{ones}(1,k)/ B(I,:);$\\
$a(I) = x;$\\
$A = [A;a];$
}
\caption{Algorithm to compute $A$}\label{alg:geta}
\end{algorithm}

Based on Lemma \ref{lem:getA}, to obtain a scheme $(A,B)$ robust to \textbf{any} $s$ stragglers, we only need to furnish a $B$ satisfying Condition~\ref{cond:bspan}. A trivial $B$ that works is $B = \mathbf{1}_{n\times k}$, the all ones matrix. However, this is wasteful since it implies that each worker gets all the partitions and computes the full gradient. Our goal is to construct $B$ satisfying Condition~\ref{cond:bspan} while also being as sparse as possible in each row. In this regard, we have the following theorem, which gives a lower bound on the number of non-zeros in any row of $B$.
\begin{theorem}[Lower Bound on B's density]\label{lem:lbB}
Consider any scheme $(A,B)$ robust to \textbf{any} $s$ stragglers, given $n$ workers (with $s<n$) and $k$ partitions. Then, if all rows of $B$ have the same number of non-zeros, we must have: $\norm{b_i}{0}\geq \frac{k}{n}(s+1)$ for any $i\in [n]$.
\end{theorem}
Theorem~\ref{lem:lbB} implies that any scheme $(A,B)$ that assigns the same amount of data to all the workers must assign at least $\frac{s+1}{n}$ fraction of the data to each worker. Since this fraction is independent of $k$, for the remainder of this paper we shall assume that $k=n$ \text{i.e.} the number of partitions is the same as the number of workers. In this case, we want $B$ to be a square matrix satisfying Condition \ref{cond:bspan}, with each row having at least $(s+1)$ non-zeros. In the sequel, we demonstrate two constructions for $B$ which satisfy Condition \ref{cond:bspan} and achieve the density lower bound.

\subsection{Fractional Repetition Scheme} \label{sec:fractional}
In this section, we provide a construction for $B$ that works by replicating the task done by a subset of the workers. We note that this construction is only applicable when the number of workers, $n$, is a multiple of $(s+1)$, where $s$ is the number of stragglers we seek tolerance to. In this case, the construction is as follows:
\begin{itemize}
\item We divide the $n$ workers into $(s+1)$ groups of size $(n/(s+1))$.\vspace*{-0.1cm}
\item In each group, we divide all the data equally and disjointly, assigning $(s+1)$ partitions to each worker\vspace*{-0.1cm}
\item All the groups are replicas of each other\vspace*{-0.1cm}
\item When finished computing, every worker transmits the sum of its partial gradients\vspace*{-0.1cm}
\end{itemize}
\begin{figure}[h]
\centering
\begin{subfigure}{0.8\textwidth}\hspace*{1in}
\includegraphics[scale=0.3]{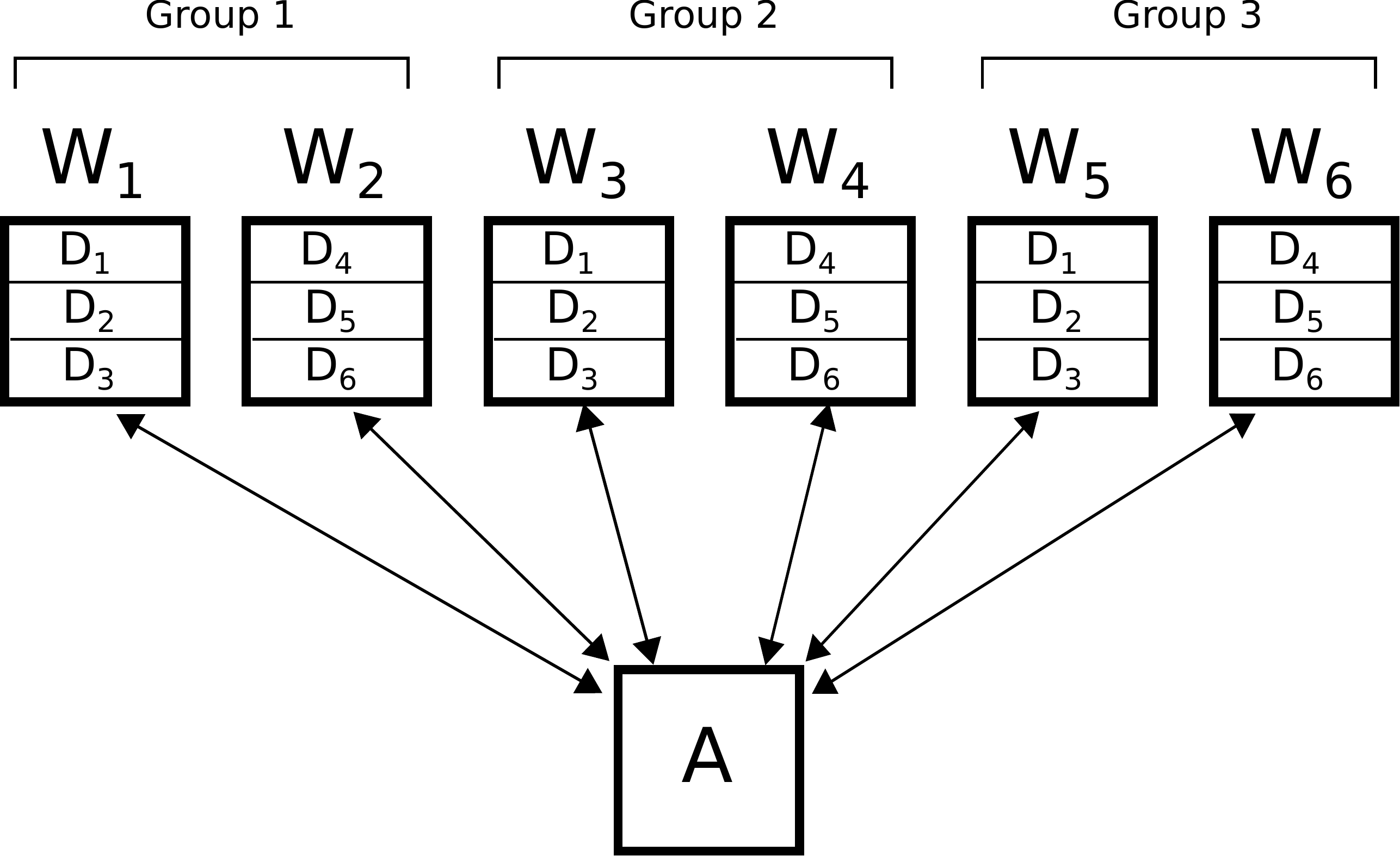}
\end{subfigure}
\caption{Fractional Repetition Scheme for $n=6,s=2$}\label{fig:rep}
\end{figure}
Fig. \ref{fig:rep} shows an instance of the above construction for $n=6,s=2$. A general description of $B$ constructed in this way (denoted as $B_{frac}$) is shown in Eq. \eqref{eq:bfrac}. 
Each group of workers in this scheme can be denoted by a block matrix $\overline{B}_{\text{block}}(n,s) \in \real^{\frac{n}{s+1}\times n}$. We define:
\begin{align}
\overline{B}_{\text{block}}(n,s) = \begin{bmatrix}\;
\mathbf{1}_{1\times (s+1)} & \mathbf{0}_{1\times (s+1)} & \cdots \,\cdots & \mathbf{0}_{1\times (s+1)} \\
\mathbf{0}_{1\times (s+1)} & \mathbf{1}_{1\times (s+1)} & \cdots \,\cdots & \mathbf{0}_{1\times (s+1)} \\
\vdots & \vdots &  \ddots & \vdots \\
\mathbf{0}_{1\times (s+1)} & \mathbf{0}_{1\times (s+1)} & \cdots \,\cdots & \mathbf{1}_{1\times (s+1)}
\;\end{bmatrix}_{\frac{n}{s+1}\times n}
\end{align}
Thus, the first worker in the group gets the first $(s+1)$ partitions, the second worker gets the second $(s+1)$ partitions, and so on. Then, $B$ is simply $(s+1)$ replicated copies of $\overline{B}_{\text{block}}(n,s)$:
\begin{equation}\label{eq:bfrac}
B = B_{frac} = \begin{bmatrix}\;
\overline{B}_{\text{block}}^{(1)}\\

\overline{B}_{\text{block}}^{(2)}\\

\vdots\\

\overline{B}_{\text{block}}^{(s+1)}

\;\end{bmatrix}_{n\times n}
\end{equation}
where for each $t\in \{1, \ldots, s+1\}$, $\overline{B}_{\text{block}}^{(t)} = \overline{B}_{\text{block}}(n,s)$.\\

It is easy to see that this construction can yield robustness to \textbf{any} $s$ stragglers. Since any particular partition of data is replicated over $(s+1)$ workers, any $s$ stragglers would leave at least one non-straggler worker to process it. We have the following theorem.
\begin{theorem}\label{thm:frac}
Consider $B_{frac}$ constructed as in Eq.~\eqref{eq:bfrac}, for a given number of workers $n$ and stragglers $s(<n)$. Then, $B_{frac}$ satisfies the B-Span condition (Condition \ref{cond:bspan}). Consequently, the scheme $(A,B_{frac})$, with $A$ constructed using Algorithm \ref{alg:geta}, is robust to any $s$ stragglers. 
\end{theorem}
The construction of $B_{frac}$ matches the density lower bound in Theorem~\ref{lem:lbB} and, the above theorem shows that the scheme $(A,B_{frac})$, with $A$ constructed from Algorithm \ref{alg:geta}, is robust to $s$ stragglers.
\subsection{Cyclic Repetition Scheme}\label{sec:cyclic}
In this section we provide an alternate construction for $B$ which also matches the lower bound in Theorem~\ref{lem:lbB} and satisfies Condition~\ref{cond:bspan}. However, in contrast to construction in the previous section, this construction does not require $n$ to be divisible by $(s+1)$. Here, instead of assigning disjoint collections of partitions, we consider a cyclic assignment of $(s+1)$ partitions to the workers. We construct a $B = B_{cyc}$ with the following support structure:
\begin{align}\label{eq:bsupp}
&\vspace*{0.2cm}\nonumber\\
&\supp(B_{cyc}) = \begin{bmatrix}\;
\bovermat{s+1}{\star & \star & \cdots & \star & \star & 0} & 0 & \cdots & 0 & 0\\
0 & \star & \star & \cdots & \star & \star & 0 & \cdots & 0 & 0\\
\vdots & \vdots & \vdots & \vdots & \vdots & \vdots & \ddots & \ddots & \vdots & \vdots\\
0 & 0 & \cdots & 0 & 0 & \star & \star & \cdots & \star & \star\\
\vdots & \vdots & \vdots & \vdots & \vdots & \vdots & \ddots & \ddots & \vdots & \vdots\\
\star & \cdots & \star & \star & 0 & 0 & \cdots & 0 & 0 & \star 
\;\end{bmatrix}_{n\times n}
\end{align}
where $\star$ indicates non-zero entries in $B_{cyc}$. So, the first row of $B_{cyc}$ has its first $(s+1)$ entries assigned as non-zero. As we move down the rows, the positions of the $(s+1)$ non-zero entries shift one step to the right, and cycle around until the last row.\\

Given the support structure in Eq.~\ref{eq:bsupp}, the actual non-zero entries must be carefully assigned in order to satisfy Condition~\ref{cond:bspan}. The basic idea is to pick every row of $B_{cyc}$, with its particular support, to lie in a suitable subspace $S$ that contains the all ones vector $\mathbf{1}_{n\times 1}$. We consider a $(n-s)$ dimensional subspace, $S = \{x\in \real^n \,\vert\, Hx = 0, \, H\in \real^{s\times n}\}$ \textit{i.e.} the null space of the matrix $H$, for some $H$ satisfying $H\mathbf{1}=0$. Now, to make the rows of $B_{cyc}$ lie in $S$, we require that the null space of $H$ must contain vectors with all the different supports in Eq.~\ref{eq:bsupp}. This turns out to be equivalent to requiring that any $s$ columns of $H$ are linearly independent, and is also referred to as the MDS property in coding theory. We show that a random choice of $H$ suffices for this, and we are able to construct a $B_{cyc}$ with the support structure in Eq.~\ref{eq:bsupp}. Moreover, for any $(n-s)$ rows of $B_{cyc}$, we show that their linear span also contains $\mathbf{1}_{n\times 1}$, thereby satisfying Condition~\ref{cond:bspan}. Algorithm~\ref{alg:consb} describes the construction of $B_{cyc}$ (in MATLAB syntax) and, we have the following theorem.
\begin{algorithm}[t]
			\SetAlgoLined
			\SetKwInOut{Input}{Input}\SetKwInOut{Output}{Output}
			\Input{$n$, $s (<n)$ }
			\Output{$B \in \real^{n\times n}$ with $(s+1)$ non-zeros in each row}
			\BlankLine
			$H = \textrm{randn}(s,n);$\\
			$H(:,n) = -\textrm{sum}(H(:,1:n-1),2);$\\
			$B = \textrm{zeros}(n);$\\
			\For{$i=1:n$}{
				$j = \textrm{mod}(i-1:s+i-1,n)+1;$\\
				$B(i,j) = [1; -H(:,j(2:s+1)) \backslash H(:,j(1))];$
			}
		\caption{Algorithm to construct $B = B_{cyc}$}\label{alg:consb}
\end{algorithm}

\begin{theorem}\label{thm:cyc}
Consider $B_{cyc}$ constructed using the randomized construction in Algorithm~\ref{alg:consb}, for a given number of workers $n$ and stragglers $s(<n)$. Then, with probability $1$, $B_{cyc}$ satisfies the B-Span condition (Condition~\ref{cond:bspan}). Consequently, the scheme $(A,B_{cyc})$, with $A$ constructed using Algorithm \ref{alg:geta}, is robust to any $s$ stragglers.
\end{theorem}
\section{Partial Stragglers}\label{sec:partstragg}
\begin{figure}[b]
\centering
\begin{subfigure}{0.4\textwidth}
\includegraphics[scale=0.3]{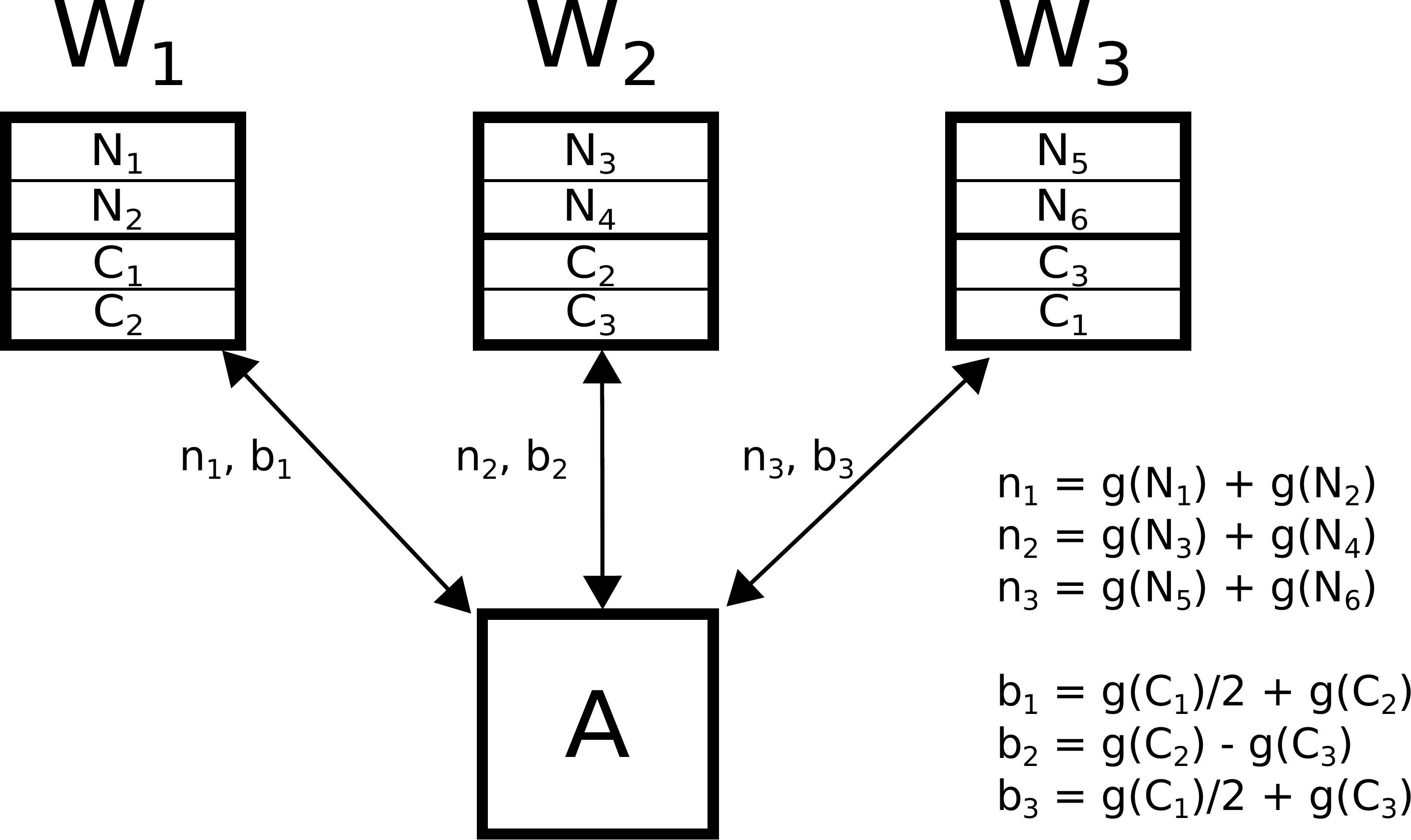}
\end{subfigure}
\caption{Scheme for Partial Stragglers, $n=3,s=1, \alpha=2$. $g(\cdot)$ represents the partial gradient.}\label{fig:part}
\vspace*{-0.5cm}
\end{figure}
In this section, we revisit our earlier assumption of \textit{full stragglers}. Under a \textit{full straggler} assumption, Theorem~\ref{lem:lbB} shows that any non-straggler worker must incur an $(s+1)$-factor overhead in computation, if we want to attain tolerance to any $s$ stragglers. This may be prohibitively huge in many situations. One way to mitigate this is by allowing at least some work to be done also by the straggling workers. Therefore, in this section, we consider a more plausible scenario of slow workers, but assume a known slowdown factor. We say that a straggler is an $\alpha$-\textit{partial straggler} (with $\alpha > 1 $) if it is at most $\alpha$ slower than any non-straggler. This means that if a non-straggler completes a task in time $T$, an $\alpha$-\textit{partial straggler} would require at most $\alpha T$ time to complete it. Now, we augment our previous schemes (in Section \ref{sec:fractional} and Section \ref{sec:cyclic}) to be robust to \textbf{any} $s$ stragglers, assuming that any straggler is an $\alpha$-\textit{partial straggler}.\\

Note that our earlier constructions are still applicable: a scheme $(A,B)$, with $B=B_{frac}$ or $B = B_{cyc}$, would still provide robustness to $s$ partial stragglers. However, given that no machine is slower than a factor of $\alpha$, a more efficient scheme is possible by exploiting at least some computation on every machine. Our basic idea is to couple our earlier schemes with a naive distribution scheme, but on different parts of the data. We split the data into a \textit{naive} component, and a \textit{coded} component. The key is to do the split such that whenever an $\alpha$-partial straggler is done processing its \textit{naive} partitions, a non-straggler would be done processing both its \textit{naive} and \textit{coded} partitions.\\

In general, for any $(n, s, \alpha)$, our two-stage scheme works as follows:
\begin{itemize}
\item We split the data $\mathbf{D}$ into $n + n\frac{s+1}{\alpha-1}$ equal-sized partitions --- of which $n$ partitions are \textit{coded} components, and the rest are \textit{naive} components
\item Each worker gets $\frac{s+1}{\alpha-1}$ \textit{naive} partitions, distributed disjointly.
\item Each worker gets $(s+1)$ \textit{coded} partitions, distributed according to an $(A,B)$ distribution scheme robust to $s$ stragglers (e.g. with $B=B_{frac}$ or $B=B_{cyc}$)
\item Any worker, $W_i$, first processes all its \textit{naive} partitions and sends the sum of their gradients to the aggregator. It then processes its \textit{coded} partitions, and sends a linear combination, as per the $(A,B)$ distribution scheme.
\end{itemize}

Note that each worker now has to send two partial gradients (instead of one, as in earlier schemes). However, a speedup gained in processing a smaller fraction of the data may mitigate this overhead in communication, since each non-straggler only has to process a $\frac{s+1}{n}\left(\frac{\alpha}{s+\alpha}\right)$ fraction of the data, as opposed to a $\frac{s+1}{n}$ fraction in \textit{full straggler} schemes. Thus, when computation is the bottleneck, adopting a partial stragglers scheme may not hurt the overall efficiency. On the other hand, when communication is the bottleneck (and if a $2\times$ overhead is prohibitive), a full straggler scheme may be a better choice even with its (s+1)-factor overhead in computation for the non-straggler workers.\\

Fig. \ref{fig:part} illustrates our two-stage strategy for $n=3, s=1, \alpha=2$. We see that each non-straggler gets $4/9 = 0.44$ fraction of the data, instead of a $2/3 = 0.67$ fraction (for e.g. in Fig \ref{fig:coded}).
\section{Experiments}\label{sec:expts}
\begin{figure*}[t]
\centering
\begin{subfigure}{0.8\textwidth}
\includegraphics[scale=0.3]{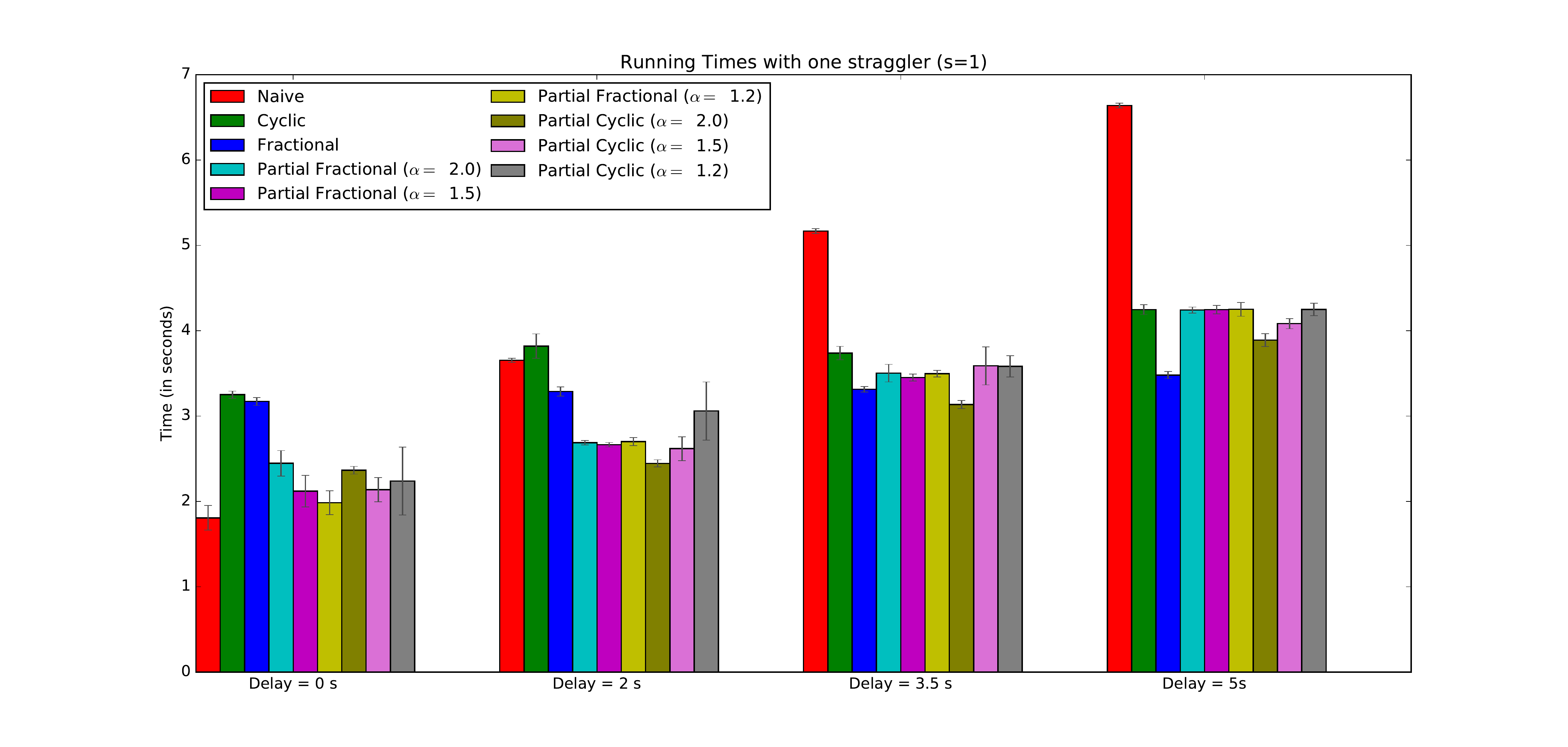}
\caption{$s=1$ Straggler}
\label{fig:del1}
\end{subfigure}
\begin{subfigure}{0.8\textwidth}
\includegraphics[scale=0.3]{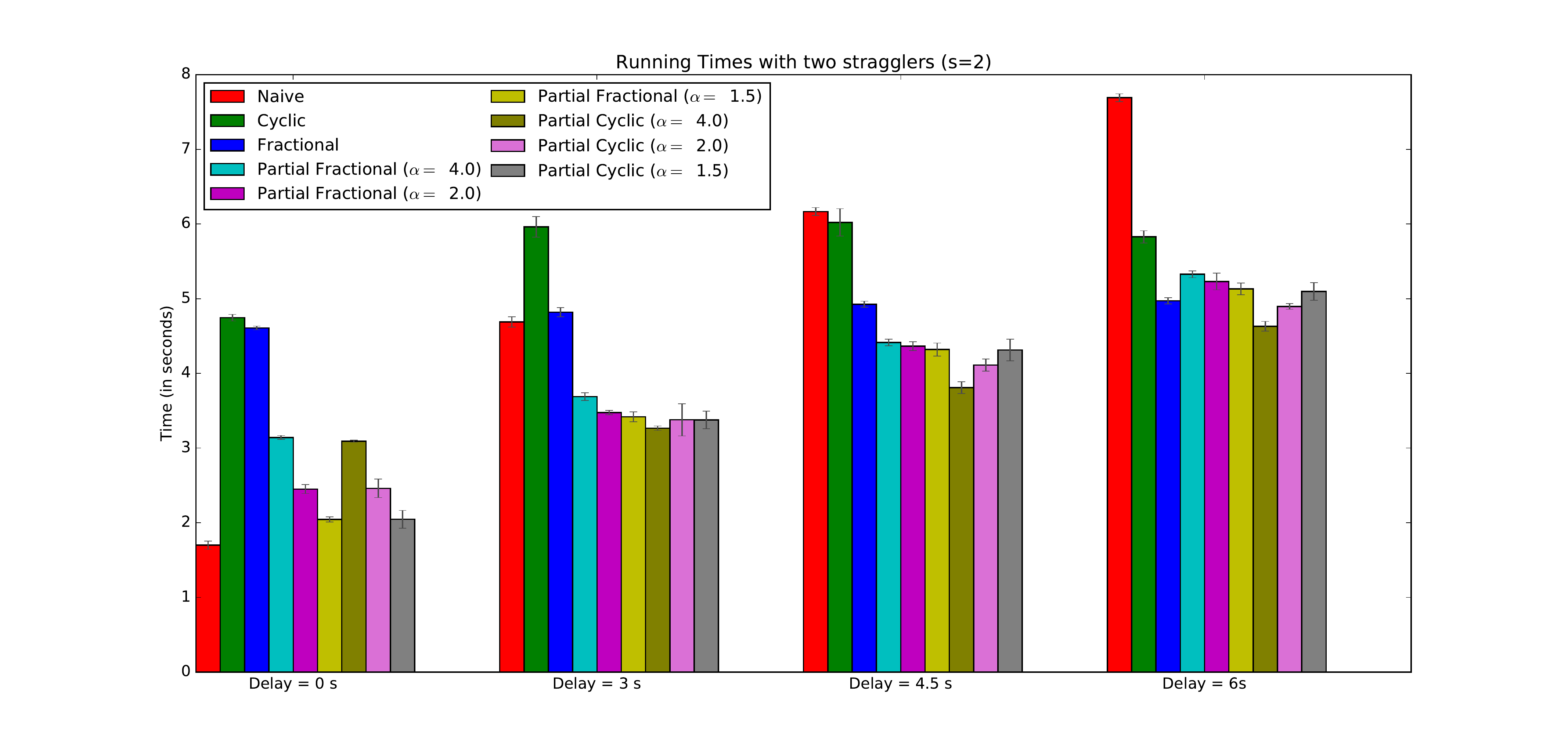}
\caption{$s=2$ Stragglers}
\label{fig:del2}
\end{subfigure}
\caption{Empirical running times on Amazon EC2 with $n=12$ machines for $s=1$ and $s=2$ stragglers. In this experiment, the stragglers are artificially delayed while the other machines run at normal speed. We note that the partial straggler schemes have much lower data replication, for example with $\alpha=1.2$ we need to only replicate approximately $10\%$ of the data.}
\label{fig:delays}
\end{figure*}
\begin{figure*}[t]
\centering\hspace*{-1cm}
\begin{subfigure}{0.33\textwidth}
\includegraphics[scale=0.33]{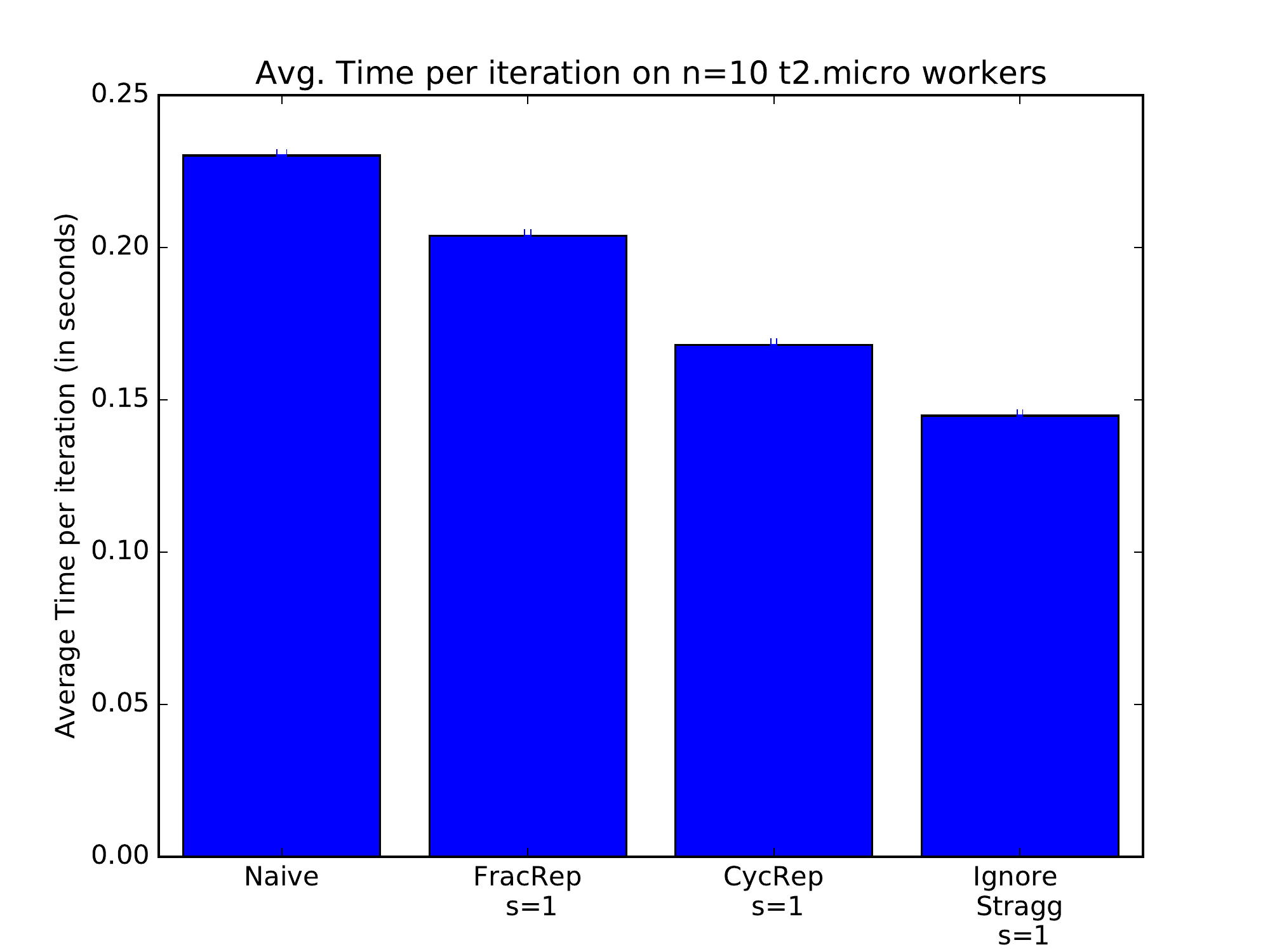}
\end{subfigure}\hspace*{0.5cm}
\begin{subfigure}{0.33\textwidth}
\includegraphics[scale=0.33]{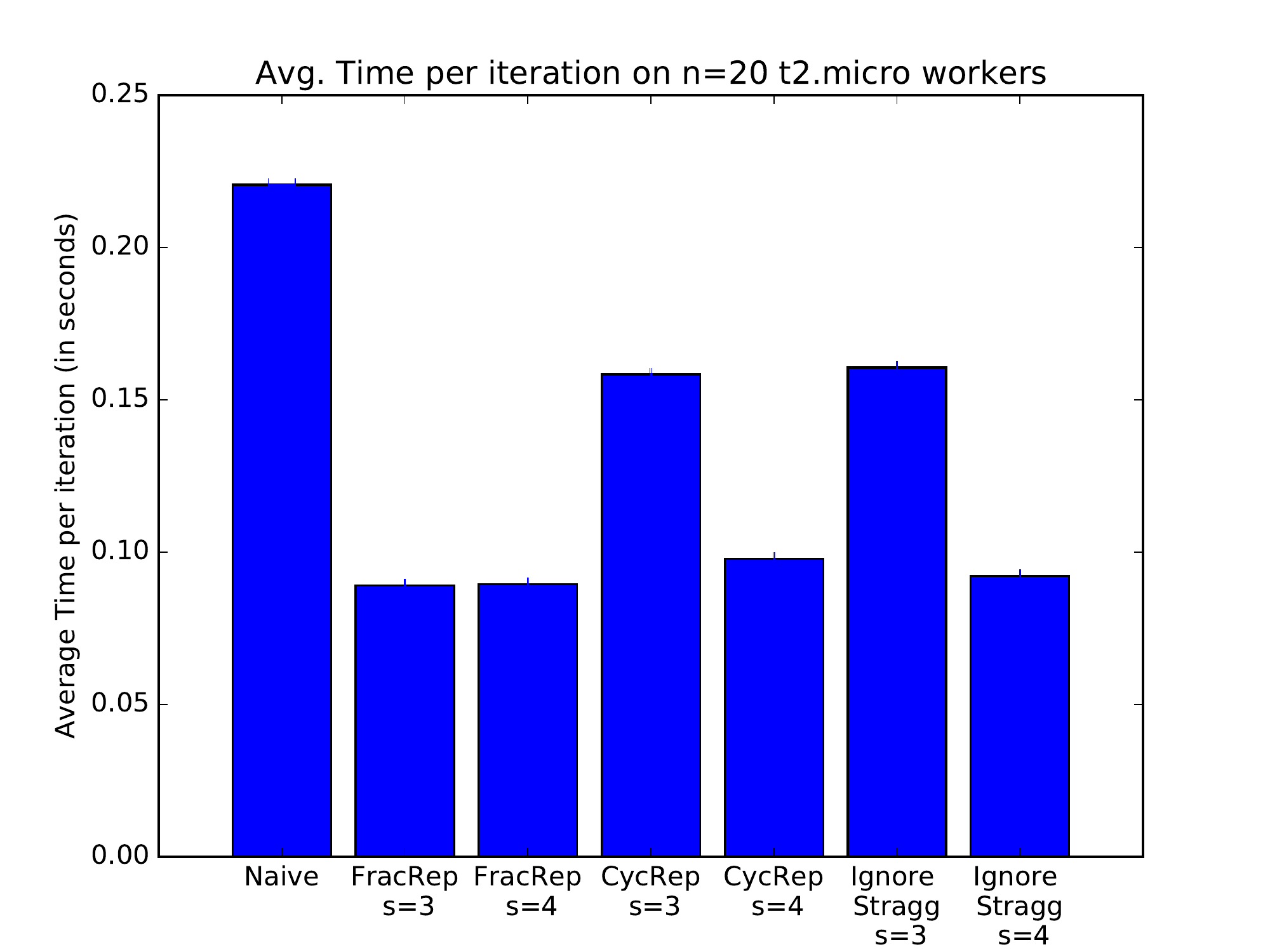}
\end{subfigure}\hspace*{0.5cm}
\begin{subfigure}{0.33\textwidth}
\includegraphics[scale=0.33]{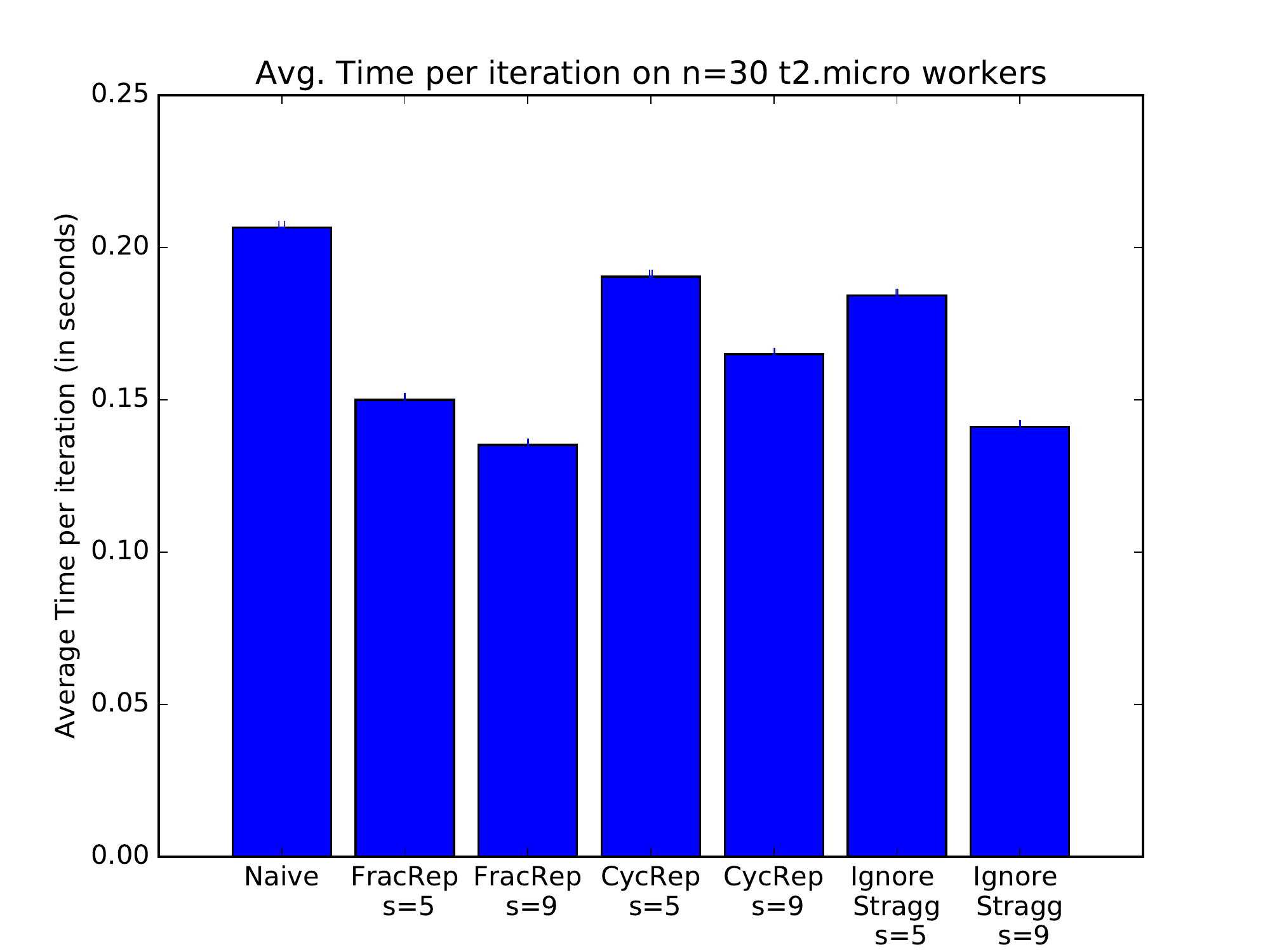}
\end{subfigure}
\caption{Avg. Time per iteration on Amazon Employee Access dataset.}\label{fig:exp1}
\end{figure*}
In this section, we present experimental results on Amazon EC2, comparing our proposed gradient coding schemes with baseline approaches. We compare our approaches against: $(1)$ the \textit{naive} scheme, where the data is divided uniformly across all workers without replication and the aggregator waits for all workers to send their gradients, and $(2)$ the \textit{ignoring $s$ stragglers} scheme where the data is divided as in the naive scheme, however the aggregator performs an update step after any $n-s$ workers have successfully sent their gradient.
\begin{figure*}[t]
\centering\hspace*{-1cm}
\begin{subfigure}{0.33\textwidth}
\includegraphics[scale=0.33]{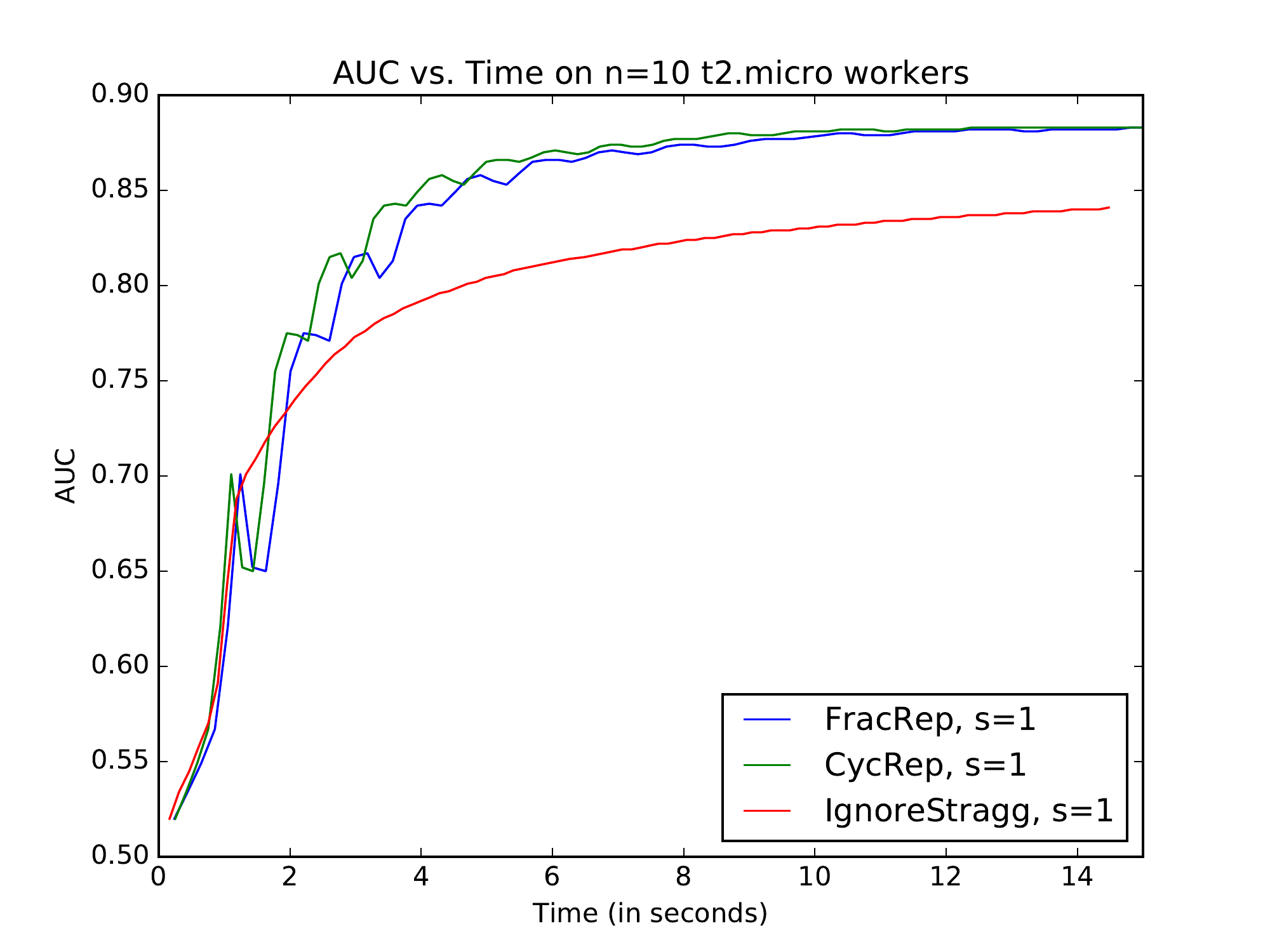}
\end{subfigure}\hspace*{0.5cm}
\begin{subfigure}{0.33\textwidth}
\includegraphics[scale=0.33]{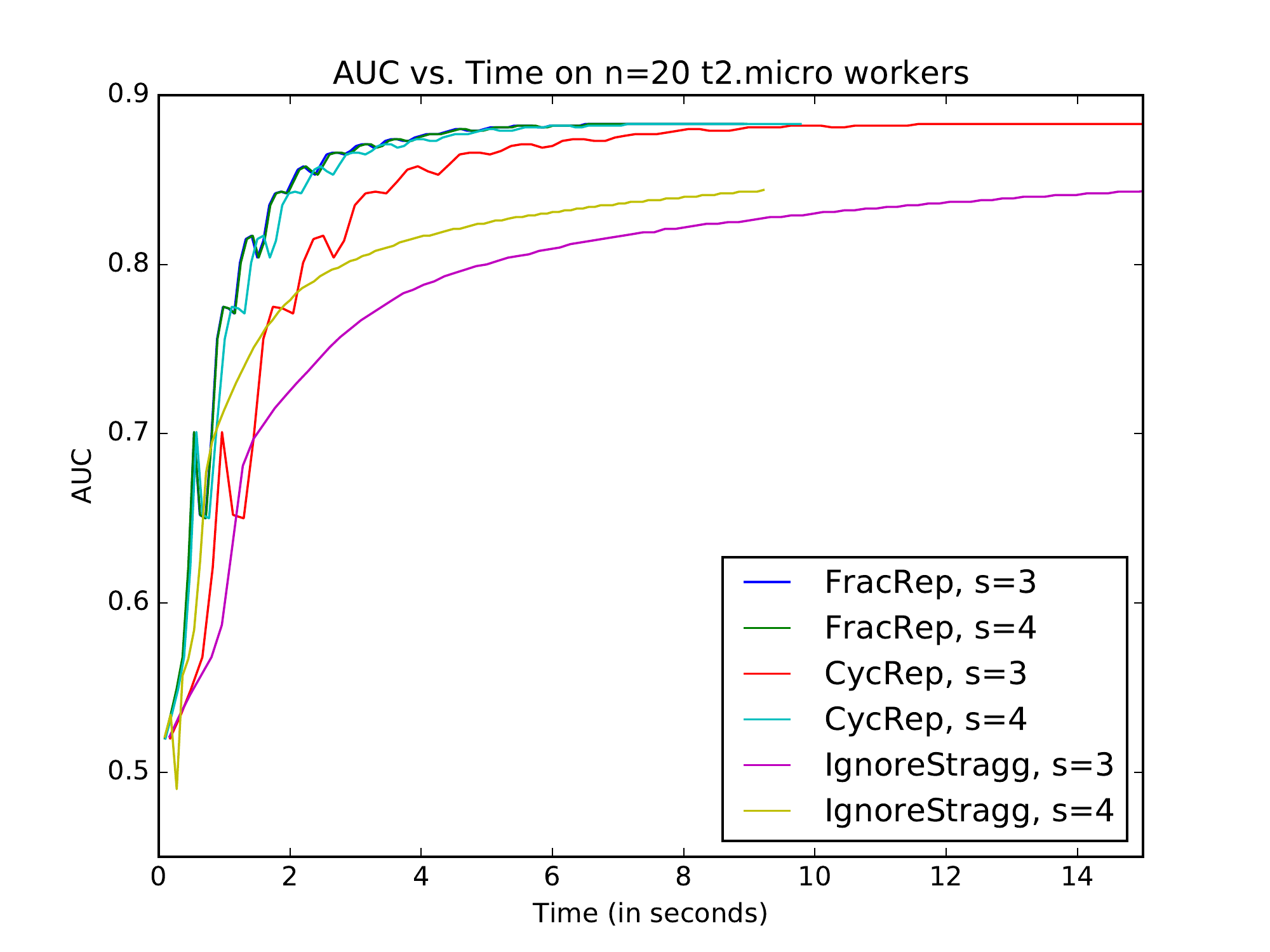}
\end{subfigure}\hspace*{0.5cm}
\begin{subfigure}{0.33\textwidth}
\includegraphics[scale=0.33]{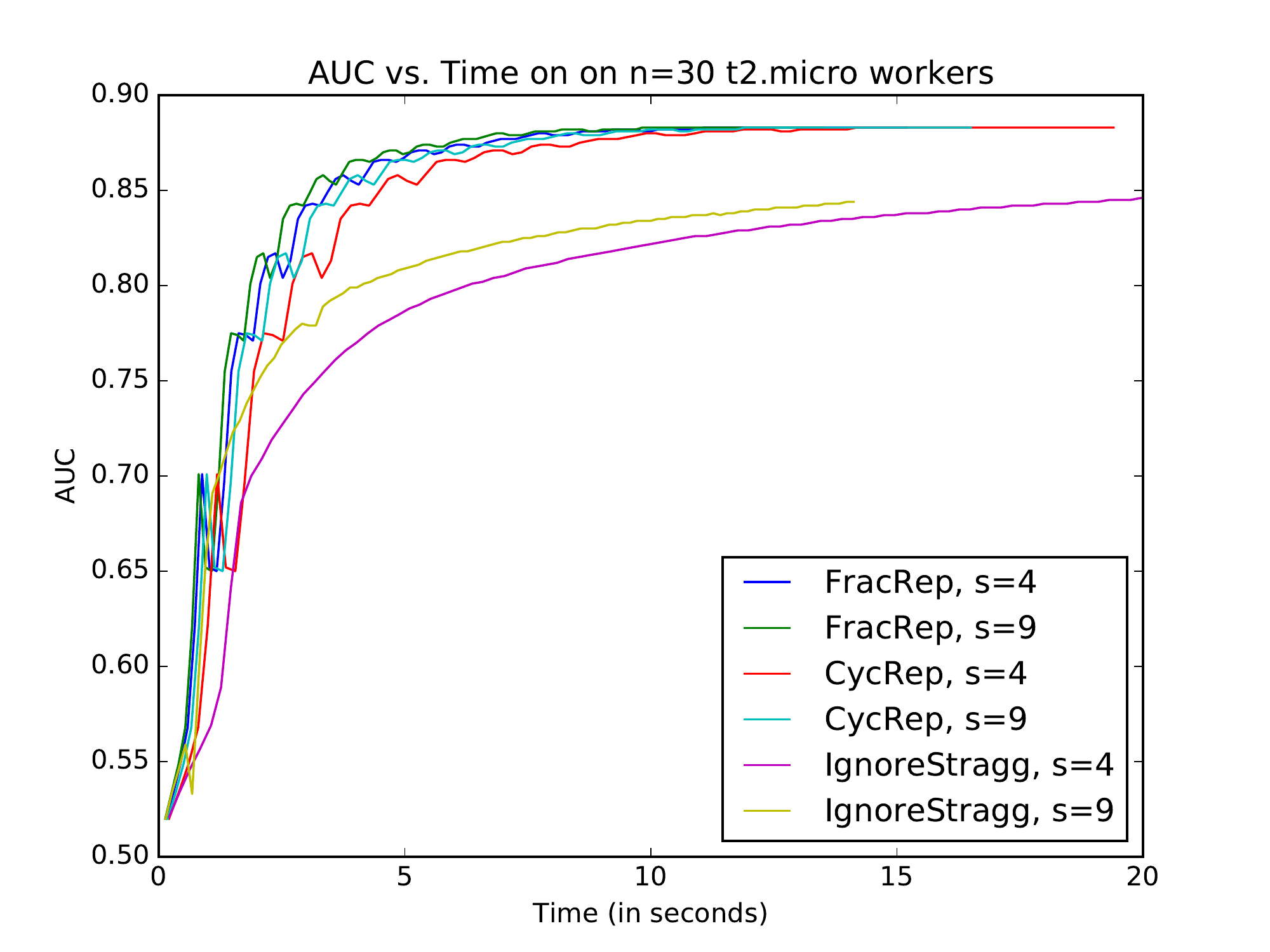}
\end{subfigure}
\caption{AUC vs Time on Amazon Employee Access dataset. 
The two proposed methods are FracRep and CycRep compared against the frequently used approach of Ignoring $s$ stragglers. As can be seen, gradient coding achieves significantly better generalization error on a true holdout.}\label{fig:exp2}
\end{figure*}
\subsection{Experimental setup}
We implemented all methods in python using MPI4py \citep{Dalcin}, an open source MPI implementation. Based on the method being considered, each worker loads a certain number of partitions of the data into memory before starting the iterations. In iteration $t$ the aggregator sends the latest model $\beta^{(t)}$ to all the workers (using \texttt{Isend()}). Each worker receives the model (using \texttt{Irecv()}) and starts a gradient computation. Once finished, it sends its gradient(s) back to the aggregator. When sufficiently many workers have returned with their gradients, the aggregator computes the overall gradient, performs a descent step, and moves on to the next iteration.\\

Our experiments were performed using two different worker instance types on Amazon EC2: \texttt{m1.small} and \texttt{t2.micro} --- these are very small, very low-cost EC2 instances. We also observed that our system was often bottlenecked by the number of incoming connections \textit{i.e.} all workers trying to talk to the master concurrently. For that reason, and to mitigate this additional overhead to some degree, we used a larger master instance of \texttt{c3.8xlarge} in our experiments. \\

We ran the various approaches to train logistic regression models, a well-understood convex problem that is widely used in practice. Moreover, Logistic regression models are often expanded by including interaction terms that are often one-hot encoded for categorical features. This can lead to 100's of thousands of parameters (or more) in the trained models. To train the logistic regression models for using our proposed scheme (or the \textit{naive} scheme), we used Nesterov's Accelerated Gradient descent with a constant learning rate, where the constant was chosen optimally from a range. Note that other optimizers such as LBFGS would have also been applicable here since we obtain the full gradient in our schemes. For the \textit{ignoring s stragglers} approach, we used gradient descent with a learning rate of $c_1/(t+c_2)$ (which is typical for SGD), where $c_1$ and $c_2$ were also chosen optimally in a range. We did not use NAG here since it is unstable to noisy gradients. While we do not present any empirical results, we refer the reader to \cite{devolder2014first} for a theoretical and empirical analysis of the effect of noisy gradients in NAG. Thus another advantage of our schemes over \textit{ignoring s stragglers} is that the latter cannot be combined with NAG because errors may quickly accumulate and eventually cause the method to diverge.
\subsection{Results}
\textbf{Artificial Dataset:} In our first experiment, we solved a logistic regression problem on a artificially generated dataset. We generated a dataset of $d = 554400$ samples $\mathbf{D} = \{(x_1,y_1),\ldots, (x_d, y_d)\}$, using the model $x \sim 0.5\times \mathcal{N}(\mu_1,I) + 0.5\times\mathcal{N}(\mu_2,I)$ (for random $\mu_1,\mu_2\in \real^{p}$), and $y \sim Ber(\kappa)$, with $\kappa = 1/ (\exp(2 x^T\beta^*) + 1)$, where $\beta^*\in \real^{p}$ is the true regressor. In our experiments, we used a model dimension of $p = 100$, and chose $\beta^*$ randomly.\\

In this experiment, we also artificially added delays to $s$ random workers in each iteration (using \texttt{time.sleep()}). Figure \ref{fig:delays} presents the results of our experiments with $s=1$ and $s=2$ stragglers, on a cluster of $n=12$ \texttt{m1.small} machines. As expected, the baseline \textit{naive} scheme that waits for the stragglers has poorer performance as the delay increases. The \textit{Cyclic} and \textit{Fractional} schemes were designed for one straggler in Figure \ref{fig:del1} and for two stragglers in Figure \ref{fig:del2}. Therefore, we expect that these two schemes would not be influenced at all by the delay of the stragglers (up to some variance due to implementation overheads). The \textit{partial straggler} schemes were designed for various $\alpha$.  Recall that for partial straggler schemes, $\alpha$ denotes the \textit{slowdown} factor.\\

\textbf{Real Dataset:} Next, we trained a logistic regression model on the Amazon Employee Access dataset from Kaggle \footnote{https://www.kaggle.com/c/amazon-employee-access-challenge}. We used $d=26200$ training samples, and a model dimension of $p=241915$ (after one-hot encoding with interaction terms). These experiments were run on $n=10,20,30$ \texttt{t2.micro} instances on Amazon EC2.\\

In Figure~\ref{fig:exp2} we show the Generalization AUC of our method (FracRep and CycRep) versus \textit{ignoring $s$ stragglers} (IgnoreStragg). As can be seen, Gradient coding achieved significantly better generalization error. We emphasize that the results in figures \ref{fig:exp1} and \ref{fig:exp2} do not use any artificial straggling, only the natural delays introduced by the EC2 cluster.\\

How is this stark difference possible? When stragglers were ignored we were, at best, receiving a stochastic gradient (when random machines are straggling in each iteration). As alluded to earlier, in this case the best we could do as an optimization algorithm is to run gradient descent as it is robust to noise. When using gradient coding however, we could retrieve the full gradient which gave us access to faster optimization algorithms. In Figure~\ref{fig:exp2} we used Nesterov's Accelerated Gradient (NAG).\\

Another advantage of using full gradients is that we can guarantee that we are training on the same distribution as the one the training set was drawn from. This is not true for the approach that ignores stragglers. If a particular machine is more likely to be a straggler, samples on that machine will likely be underrepresented in the final model,  unless particular countermeasures are deployed. There may even be inherent reasons why a particular sample will systematically be excluded when we ignore stragglers. For example, in structured models such as linear-chain CRFs, the computation of the gradient is proportional to the length of the sequence. Therefore, extraordinarily long examples can be ignored very frequently. 
\section{Conclusion}
In this paper, we have experimented with various \textit{gradient coding} ideas on Amazon EC2 instances. This is a complex trade-off space between model sizes, number of samples, worker configurations, and number of workers. Our proposed schemes create computation overheads while keeping communication the same.\\

The benefit of this additional computation is fault-tolerance: we are able to recover full gradients, even if $s$ machines do not deliver their assigned work, or are slow in doing so. Moreover, our partial straggler schemes provide fault tolerance while allowing all machines to do partial work. They however require an extra round of communication. An interesting open problem here is whether partial work on all machines is possible without this extra round of communication. Another open question under our framework is that of approximate gradient coding: can we get a vector that \textit{is close to the true gradient}, with lesser computation overheads ? Ignoring stragglers does give the approximate gradient in a sense. However, is it possible to have a better approximation with on little computation overheads (relative to \textit{gradient coding}) ?\\

For several model-cluster configurations that we tested, communication was the bottleneck and hence the additional computation's effect on iteration times was negligible. This is the regime where gradient coding is most useful. However, this design space needs further exploration, that is also varying as different architectures change the parameter landscape. Overall, we believe that gradient coding is an interesting idea to add in the distributed large-scale learning arsenal. 

\section*{Acknowledgements}
This research has been supported by NSF Grants CCF 1344364, 1407278, 1422549, 1618689 and ARO YIP W911NF-14-1-0258.
\newpage
\bibliography{stragg}
\bibliographystyle{apa}
\newpage
\onecolumn
\section{Appendix - Proofs}
\subsection{Proof of Lemma \ref{lem:getA}}
By Condition \ref{cond:bspan}, we know that for any $I\subseteq [n],\, \lvert I \rvert = n-s$, we have $\mathbf{1}\in \text{span}\{b_i\,\vert\, i \in I\}$. In other words, there exists at least one $x\in \real^{(n-s)}$ such that:
\begin{equation}
x B(I,:) = \mathbf{1}
\end{equation}
Therefore, by construction, we have: $AB = \mathbf{1}_{\binom{n}{s}\times n}$, and the scheme $(A,B)$ is robust to \textbf{any} $s$ stragglers.

\subsection{Proof of Theorem \ref{lem:lbB}}
Consider any scheme $(A,B)$ robust to \textbf{any} $s$ stragglers, with $B\in \real^{n\times k}$. Now, construct a bipartite graph between $n$ workers, $\{W_1,\ldots,W_n\}$, and $k$ partitions, $\{P_1,\ldots,P_k\}$, where we add an edge $(i,j)$ if worker $i$ and partition $j$ is worker $i$ has access to partition $j$. In other words, for any $i\in [n], j\in [k]$: 
\begin{equation}
e_{ij} = \begin{cases}
1 & \,\text{if } B(i,j)\ne 0\\
0 & \,\text{otherwise}
\end{cases}
\end{equation}
Now, it is easy to see that the degree of the $i^{th}$ worker $W_i$ is $\norm{b_i}{0}$.

Also, for any partition $P_j$, its degree must be at least $(s+1)$. If its degree is $s$ or less, then consider the scenario where all its neighbors are stragglers. In this case, there is no non-straggler worker with access to $P_j$, which contradicts robustness to \textbf{any} $s$ stragglers. 

Based on the above discussion, and using the fact that the sum of degrees of the workers in the bipartite graph must be the same as the sum of degrees of partitions, we get:
\begin{equation}
\sum_{i=1}^n \norm{b_i}{0} \geq k(s+1)
\end{equation}
Since we assume all workers get access to the same number of partitions, this gives:
\begin{equation}
\norm{b_i}{0}\geq \frac{k(s+1)}{n},\, \text{ for any } i\in [n]
\end{equation}

\subsection{Proof of Theorem \ref{thm:frac}}
Consider groups of partitions $\{G_1, \ldots, G_{n/(s+1)}\}$ as follows:
\begin{align}
G_1 &= \{P_1, \ldots, P_{s+1}\}\nonumber\\
G_2 &= \{P_{s+2},\ldots, P_{2s+2}\}\nonumber\\
\vdots & \\
G_{n/(s+1)} &= \{P_{n-s}, \ldots, P_n\}
\end{align}

Fix some set $I\subseteq [n], \lvert I\rvert = n-s$. Based on our construction, it is easy to observe  that for any group $G_j$, there exists some index in $I$, say $i_{G_j} \in I$, such that the corresponding row in $B$, $b_{i_{G_j}}$ has all $1$s at partitions in $G_j$ and $0$s elsewhere. This is because there are $(s+1)$ rows of $B$ that correspond in this way to $G_j$ (one in each block $\overline{B}_{\text{block}}$), and so at least one would survive in the set $I$ of cardinality $(n-s)$. Now, it is trivial to see that:
\begin{equation}
\mathbf{1} \in \text{span}\{b_{i_{G_j}} \,\vert\, j=1, \ldots, n/(s+1)\}
\end{equation}
Also, since 
\begin{equation}
\text{span}\{b_{i_{G_j}} \,\vert\, j=1, \ldots, n/(s+1)\} \subseteq \text{span}\{b_i \,\vert\, i\in I\},
\end{equation}
we have $\mathbf{1}\in \text{span}\{b_i \,\vert\, i\in I\}$. 

Finally, since the above holds for any set $I$, we get that $B$ satisfies Condition \ref{cond:bspan}. The remainder of the theorem follows from Lemma \ref{lem:getA}.
 
\subsection{Proof of Theorem \ref{thm:cyc}}
Consider the subspace given by the null space of the random matrix $H$ (constructed in Algorithm \ref{alg:consb}):
\begin{equation}\label{eq:subS}
S = \{x \in \real^n\,\vert\, Hx = 0\}
\end{equation}
Note that $H$ has $(n-1)s$ different random values ($s$ for each column), since its last column is simply the negative sum of its previous $(n-1)$ columns. Now, we have the following Lemma listing some properties of $H$ and $S$.

\begin{lemma}\label{lem:propS}
Consider $H\in \real^{ss\times n}$ as constructed in Algorithm \ref{alg:consb}, and the subspace $S$ as defined in Eq. \ref{eq:subS}. Then, the following hold:
\begin{itemize}
\item Any $s$ columns of $H$ are linearly independent with probability $1$
\item $dim(S) = n-s$ with probability $1$
\item $\mathbf{1}\in S$, where $\mathbf{1}$ is the all-ones vector
\end{itemize}
\end{lemma}

For $i\in [n]$, let $S_i$ denote the set $S_i = \{i \mod n, (i+1) \mod n, \ldots, (i+s) \mod n\}$. Then, $S_i$ corresponds to the support of the $i^{th}$ row of $B$ in our construction, as also given by the support structure in Eq. \eqref{eq:bsupp}.

Recall that we denote the $i^{th}$ row of $B$ by $b_i$. By our construction, we have:
\begin{align}\label{eq:bi}
b_i(i) & = 1\nonumber\\
b_i(S_i\setminus \{i\}) &= -H_{S_i\setminus \{i\}}^{-1} H_i
\end{align}
Now, we have the following lemma;
\begin{lemma}\label{lem:bi}
Consider the $i^{th}$ row of $B$ constructed using Algorithm \ref{alg:consb} (also shown in Eq. \ref{eq:bi}). Then, 
\begin{itemize}
\item $b_i \in S$
\item Every element of $b_i(S_i\setminus \{i\})$ is non-zero with probability 1
\item For any subset $I\subseteq [n]$, $\lvert I\rvert =n-s$, the set of vectors $\{b_i \,\vert \, i\in I\}$ is linearly independent with probability 1
\end{itemize}
\end{lemma}

Now, using Lemma \ref{lem:bi}, we can conclude that for any subset $I\subseteq [n]$, $\lvert I\rvert =n-s$, $dim\left(\text{span}\{b_i\,\vert\, i\in I\}\right) = n-s$ and $\text{span}\{b_i\,\vert\, i\in I\}\subseteq S$. Consequently, from Lemma \ref{lem:propS}, since $dim(S) = n-s$ and $\mathbf{1}\in S$, this implies that:
\begin{equation}
\text{span}\{b_i\,\vert\, i\in I\} = S \text{ with probability 1}
\end{equation}
and, $\mathbf{1}\in \text{span}\{b_i\,\vert\, i\in I\}$. Taking union bound over every $I$ shows that $B$ satisfies Condition \ref{cond:bspan}. The remainder of the theorem follows from Lemma \ref{lem:getA}.

\subsubsection{Proof of Lemma \ref{lem:propS}}
Consider any subset $I\subseteq n$, $\lvert I\rvert = s$ such that $n \notin I$. Then, all the elements of $H_I$ are independent, and $det(H_I)$ is a polynomial in the elements of $H_I$. Consequently, since every element is drawn from a continuous probability distribution (in particular, Gaussian), the set $\{H_I\,\vert\, det(H_I)=0\}$ is a zero measure set. So, $P\left( det(H_I)\ne 0\right)=1$, and thus the columns of $H_I$ are linearly independent with probability 1.

If $n\in I$, then we have:
\begin{equation}
det(H_I) = det(\widetilde{H})
\end{equation}
where we let $\widetilde{H} = \begin{bmatrix}
H_{I\setminus \{n\}}, -\sum_{i\in [n]\setminus I} H_i
\end{bmatrix}$. The elements of $\widetilde{H}$ are independent, so using the same argument as above, we again have $P(det(H_I) = det(\widetilde{H}) \ne 0) = 1$. Finally, taking a union bound over all sets $I$ of cardinality $s$ shows that any $s$ columns of $H$ are linearly independent.

Since any $s$ columns in $H$ are linearly independent, this implies that $rank(H)=s$. Since the subspace $S$ is simply the null space of $H$, we have $dim(S) = n-s$.

Finally, since $H_n = - \sum_{i\in [n-1]} H_i$ (by construction), we have $H\mathbf{1} = 0$ and thus $\mathbf{1}\in S$.

\subsubsection{Proof of Lemma \ref{lem:bi}}
By construction of $b_i$, we have:
\begin{equation}
H b_i = H_i + H_{S_i\setminus \{i\}} b_i(S_i\setminus\{i\}) = H_i - H_i = 0
\end{equation}
Thus, $b_i\in S$.

Now, if possible, let for some $k\in S_i\setminus \{i\}$, $b_i(k)=0$. Then, since $b_i\in S$, we have:
\begin{equation}
H b_i = H_i + H_{S_i\setminus \{i,k\}} b_i(S_i\setminus\{i,k\}) = 0
\end{equation}
Consequently, the set of columns $\{j\,\vert \, j\in S_i\setminus\{i,k\}\} \cup \{i\}$ is linearly dependent which contradicts $H$ having any $s$ columns being linearly independent (in Lemma \ref{lem:propS}). Therefore, we must have every element of $b_i(S_i\setminus\{i\})$ being non-zero.

Now, consider any subset $I\subseteq [n], \lvert I\rvert = n-s$. We shall show that the matrix $B_I$ (corresponding to the rows of $B$ with indices in $I$) has rank $n-s$ with probability $1$. Consequently, the set of vectors $\{b_i \,\vert \, i\in I\}$ would be linearly independent. To show this, we consider some $n-s$ columns of $B_I$, say given by the set $J\subseteq [n], \lvert J\rvert = n-s$, and denote the sub-matrix of columns by $B_{I,J}$. Then, it suffices to show that $\text{det}(B_{I,J})\ne 0$. Now, by the construction in Algorithm \ref{alg:consb}, we have: $\text{det}(B_{I,J}) = \text{poly}_1(H)/\text{poly}_2(H)$, for some polynomials $\text{poly}_1(\cdot)$ and $\text{poly}_2(\cdot)$ in the entries of $H$. Therefore, if we can show that there exists at least one $H'$ with $H'\mathbf{1} = \mathbf{0}$ and $\text{poly}_1(H')/ \text{poly}_2(H')\ne 0$, then under a choice of i.i.d. standard Gaussian entries of $H$, we would have:
\begin{equation}
\prob\left(\text{poly}_1(H) / \text{poly}_2(H)\ne 0\right) = 1
\end{equation}

The remainder of this proof is dedicated to showing that such an $H'$ exists. To show this, we shall consider a matrix $\widetilde{B}\in \real^{n-s \times n}$ such that $\supp(\widetilde{B}) = \supp(B_I)$ and $\text{det}(\widetilde{B}_{:,J})\ne 0$, where $\widetilde{B}_{:,J}$ corresponds to the sub-matrix of $\widetilde{B}$ with columns in the set $J$. Given such a $\widetilde{B}$, we shall show that there exists an $s\times n$ matrix $H'$ (with $H' \mathbf{1} = \mathbf{0}$) such that when we run Algorithm \ref{alg:consb} with this $H'$, we get a matrix $B'$ s.t. $B'_I = \widetilde{B}$ \textit{i.e.} the output matrix from Algorithm \ref{alg:consb} is identical to our random choice $\widetilde{B}$ on the rows in the set $I$. This suffices to show the existence of an $H'$ such that $\text{poly}_1(H')/ \text{poly}_2(H')\ne 0$, since $\text{poly}_1(H')/ \text{poly}_2(H') = \text{det}(B'_{I,J}) = \text{det}(\widetilde{B}_{J})\ne 0$.

Let us pick a random matrix $\widetilde{B}$ as:
\begin{equation}
\widetilde{B} = B^{r}_I D
\end{equation}
where $B^{r}_I$ is a matrix with the same support as $B_I$ and with each non-zero entry i.i.d. standard Gaussian, and $D$ is a diagonal matrix such that $D_{ii} = \sum_{j=1}^{n-s} B_{I}^{r}(j,i), \,i\in [n]$. Note that a consequence of the above choice of $\widetilde{B}$ is that the sum of all its rows is the all $\mathbf{1}$s vector. Now, it can be shown that any $(n-s)$ columns of $\widetilde{B}$ form an invertible sub-matrix with probability 1. Let $S_i$ be the support of the $i^{th}$ row of $B$. The rows of $B^r_I$ have the supports $S_i, i\in I$. Now because of the cyclic support structure in $B$, any collection $\{i_1, i_2, \ldots, i_k\} (0 \leq k\leq n-s)$ satisfies the property:
\begin{equation}
\lvert \cup_{j=1}^{k} S_{i_j}\rvert \geq s + k
\end{equation}
Using Lemma 4 in \cite{dau13}, this implies that there is a perfect matching between the rows of $B^r_I$ and any of its $(n-s)$ columns . Consequently, with probability 1, any $(n-s)$ columns of $B^r_I$ form an invertible sub-matrix. Also, since every column of $B^r_I$ contains at least one non-zero (again, owing to the support structure of $B$), this implies that with probability 1, all the diagonal entries of $D$ are non-zero. Combining the above two observations, we can infer that any $(n-s)$ columns of $\widetilde{B}$ form an invertible sub-matrix with probability 1.

So far, we have shown existence of a matrix $\widetilde{B}$ with the following properties: (i) $\widetilde{B}$ has the same support structure as $B_I$, (ii) any $(n-s)$ columns of $\widetilde{B}$ form invertible sub-matrix, (iii) the sum of all rows of $\widetilde{B}$ is the all $\mathbf{1}$s vector. Now, for any such $\widetilde{B}$, we shall show that there exists an $H'$ such that $H'\widetilde{B}^T = \mathbf{0}$ such that any $s$ columns of $H'$ form an invertible sub-matrix. This implies that when we run Algorithm \ref{alg:consb} with this $H'$, the output matrix would be the same as $\widetilde{B}$ on the rows in the set $I$. The remainder of the proof then follows from our earlier discussion.

Now, consider any set $Q\subseteq [n], \lvert Q\rvert \leq s$. Suppose we pick any invertible $H'_{:,Q}$, and set $H'_{:,[n]\setminus Q} = - H'_{:,Q} \widetilde{B}_{:,Q}^T (\widetilde{B}_{:,[n]\setminus Q}^T)^{-1}$. Then, such an $H'$ satisfies $H'\widetilde{B}^T = 0$ and its columns in the set $Q$ form an invertible sub-matrix. Now, since invertibility on the set $Q$ simply corresponds to $\text{det}(H'_{:,Q})\ne 0$ (\textit{i.e.} some fixed polynomial being non-zero), if we actually picked a uniformly random $H'$ on the subspace $H'\widetilde{B}^{T}=0$, then
\begin{equation}
\prob\left(\text{det}(H'_{:,Q})\ne 0 \,\vert\, H'\widetilde{B}^T = 0 \right) = 1
\end{equation}
Taking a union bound over all $Q$s, we get that 
\begin{equation}
\prob\left( \text{any $s$ columns of $H'$ form an invertible sub-matrix} \,\vert\, H'\widetilde{B}^T = 0\right) = 1
\end{equation}
Thus, there exists an $H'$ satisfying $H'\widetilde{B}^T = 0$ with any $s$ of its columns forming an invertible sub-matrix. Also, since the sum of all rows of $\widetilde{B}$ is $\mathbf{1}$, this implies $H'\mathbf{1} = \mathbf{0}$.

\end{document}